\documentclass[conference]{IEEEtran}

\usepackage{graphicx}

\usepackage{amsmath,amsfonts}

\usepackage{subcaption} 
\begin{document}
%
\title{Sensing for Space Safety and Sustainability: A Deep Learning Approach with Vision Transformers}
%
%
%

\author{\IEEEauthorblockN{
Wenxuan Zhang\IEEEauthorrefmark{1} and
Peng Hu\IEEEauthorrefmark{1}\IEEEauthorrefmark{2}}
\IEEEauthorblockA{\IEEEauthorrefmark{1}Faculty of Mathematics, University of Waterloo, Waterloo, Canada}
\IEEEauthorblockA{\IEEEauthorrefmark{2}
Dept. of Electrical and Computer Engineering, University of Manitoba, Winnipeg, Canada}
{v39zhang@uwaterloo.ca, peng.hu@umanitoba.ca}
}


%
%

\markboth{Journal of \LaTeX\ Class Files,~Vol.~14, No.~8, August~2015}%
{Shell \MakeLowercase{\textit{et al.}}: Bare Demo of IEEEtran.cls for IEEE Journals}
%




\maketitle

\begin{abstract}
The rapid increase of space assets represented by small satellites in low Earth orbit can enable ubiquitous digital services for everyone. However, due to the dynamic space environment, numerous space objects, complex atmospheric conditions, and unexpected events can easily introduce adverse conditions affecting space safety, operations, and sustainability of the outer space environment. This challenge calls for responsive, effective satellite object detection (SOD) solutions that allow a small satellite to assess and respond to collision risks, with the consideration of constrained resources on a small satellite platform. This paper discusses the SOD tasks and onboard deep learning (DL) approach to the tasks. Two new DL models are proposed, called GELAN-ViT and GELAN-RepViT, which incorporate vision transformer (ViT) into the Generalized Efficient Layer Aggregation Network (GELAN) architecture and address limitations by separating the convolutional neural network and ViT paths. These models outperform the state-of-the-art YOLOv9-t in terms of mean average precision (mAP) and computational costs. On the SOD dataset, our proposed models can achieve around 95\% mAP50 with giga-floating point operations (GFLOPs) reduced by over 5.0. On the VOC 2012 dataset, they can achieve $\geq$ 60.7\% mAP50 with GFLOPs reduced by over 5.2. 
\end{abstract}


%
\IEEEpeerreviewmaketitle

\section{Introduction}
\IEEEPARstart{W}{ith} the rapid increase of advanced low-Earth-orbit (LEO) satellites in large constellations, space assets are set to play a crucial role in the future, enabling global Internet access and space relay systems for deep space missions. However, managing thousands of LEO satellites with the assurance of safety operations, space sustainability, and space situational awareness is becoming a significant challenge. For example, the increasing risk of collisions between LEO satellites and space objects can cause significant amount of space debris in various sizes and threaten the safety operations of spacecraft and outer space environment. These risks are real and will be exacerbated with the increasing space assets into the outer space. As evident from the past incidents between satellites and space debris and from the study by the Inter-Agency Space Debris Coordination Committee (IADC) \cite{IADC_study_13} indicating that ``catastrophic collisions are expected every 5 to 9 years'' in an almost perfect scenario. 

One fundamental challenge is the effective satellite object detection (SOD) for collision assessment and avoidance. In this case, a satellite, in particular, LEO satellite, should be able to efficiently detect other satellite objects at a certain field of view. Such an SOD task needs to detect the objects with high precision and low latency. In addition, considering the constrained hardware capacity and energy on a LEO satellite, the SOD task needs to be as resource efficient as possible. Currently, there is no good solution to this challenge. Despite the ease of manipulation and management, the ground radar or laser telescope facilities for satellite tracking require some tolerance of latency introduced by the contact times and the transmission of the detection results, not to mention the high cost of the facilities. The Light Detection and Ranging (LiDAR) sensors have limited detection range and are susceptible to atmospheric conditions, so the ground detection using current LiDAR sensors is hardly feasible. Therefore, to reduce the latency of SOD tasks, we need onboard solutions on the satellites so they can detect the objects individually. Such an onboard sensing is expected to conduct sensing results in near real time. Then the question comes to the sensing mechanism that can be suited for the small satellite platform, while this question has not been answered yet. 

The radar or LiDAR sensors, if we equip them on a LEO satellite, will result in significant power consumption and can take as much as 45\% of a small satellite's power budget. The vision sensors can generally provide cost-effective solutions with low power consumption. However, their effectiveness has not been fully explored for the SOD tasks. The popular object detection models, such as You Only Look Once (YOLO) based models \cite{ge2021yoloxexceedingyoloseries, wang2023yolov7, wang2024yolov9learningwantlearn}, vision transformer (ViT) based models \cite{mehta2022mobilevit, aubard2024knowledge}, focus on the regular detection tasks and how they will perform in the SOD tasks is unknown. In this paper, we aim to provide a in-depth discussion on the effectiveness of employing vision sensors with computer vision (CV) capability for SOD tasks based on the state-of-the-art and our proposed deep learning (DL) models. Specifically, our proposed models are based on ViT and the Generalized Efficient Layer Aggregation Network (GELAN) in YOLOv9 \cite{wang2024yolov9learningwantlearn} and consistently show superior performance in the SOD tasks. 

The paper is structured as follows. The related work is discussed in Section II. The proposed system model is discussed in Section III. The performance evaluation of the proposed models is presented in Section IV. The conclusive remarks are made in Section V. 

\section{Related Work}

In the domain of computer vision, object detection has been an important driving force behind many significant advancements. Convolutional neural network (CNN) based detectors are the most widely used in CV, where they are classified into one-stage and two-stage detectors. Two-stage detectors, such as R-CNN \cite{girshick2014rich}, exchange detection speed for improved accuracy by first identifying the region of interest and then making the prediction. Due to the timing requirement of SOD tasks, we should consider one-stage detectors that can usually achieve faster detection speed. Typical one-stage detectors, such as the YOLO-based models  \cite{7780460, ge2021yoloxexceedingyoloseries, wang2023yolov7, wang2024yolov9learningwantlearn}, and single-shot detection (SSD) \cite{10.1007/978-3-319-46448-0_2}, predict the bounding box and class labels simultaneously, leading to a faster detection speed. However, a study by Jiaxu \textit{et al.} \cite{Leng2019An} suggests that one-stage detectors tend to struggle with small object detection, largely due to the fixed bounding box scales. 

The YOLO-based models are known for their balance between speed and accuracy in real-time applications. YOLOv9 \cite{wang2024yolov9learningwantlearn}, builds upon the GELAN architecture and Programmable Gradient Information (PGI). GELAN enhances feature aggregation across different layers, while PGI mitigates information loss within the network. As a result, YOLOv9 is expected to excel in complex scenarios, where preserving information is key to improving prediction accuracy. YOLOv7 \cite{wang2023yolov7} excels in detection speed and accuracy for simpler tasks, but its performance may decline in complex object detection tasks due to its simpler architecture. YOLOv7 is effective in scenarios where high-speed inference is required without significantly sacrificing accuracy.

ViTs have introduced a new path in CV by capturing global dependencies within images using self-attention mechanisms \cite{Dosovitskiy2020AnII}. Unlike CNNs, which excel at local feature extraction, ViTs are adept at modelling long-range relationships of objects within an image. Models such as YOLOX-ViT \cite{aubard2024knowledge} and MobileViT \cite{mehta2022mobilevit} attempt to combine the strengths of CNNs and ViTs by integrating transformer layers into CNN architectures to improve both detection accuracy and computational efficiency.

Despite these advancements, previous approaches to integrating CNNs and ViTs, such as YOLOX-ViT and MobileViT, have limitations. These models do not separate local and global feature extraction processes, but they combine them within the same pathways. This can lead to information interference and loss, as the capacity of each layer is stretched to handle both types of features simultaneously. This issue is particularly problematic in complex detection tasks, where balancing detailed local information with a broader global context is crucial.

Our models, GELAN-ViT and GELAN-RepViT, incorporate ViT into the GELAN architecture and address limitations by separating the CNN and ViT paths, allowing each to specialize in its strengths. This mitigates the risk of information loss and optimizes feature extraction. The following section provides an overview of our system model, explaining how we separate these feature extraction processes and detailing the architecture of our proposed models.

\section{System Model}

\begin{figure*}[t]
    \centering
    \begin{subfigure}[t]{0.5\textwidth}
        \centering
        \includegraphics[width=\linewidth]{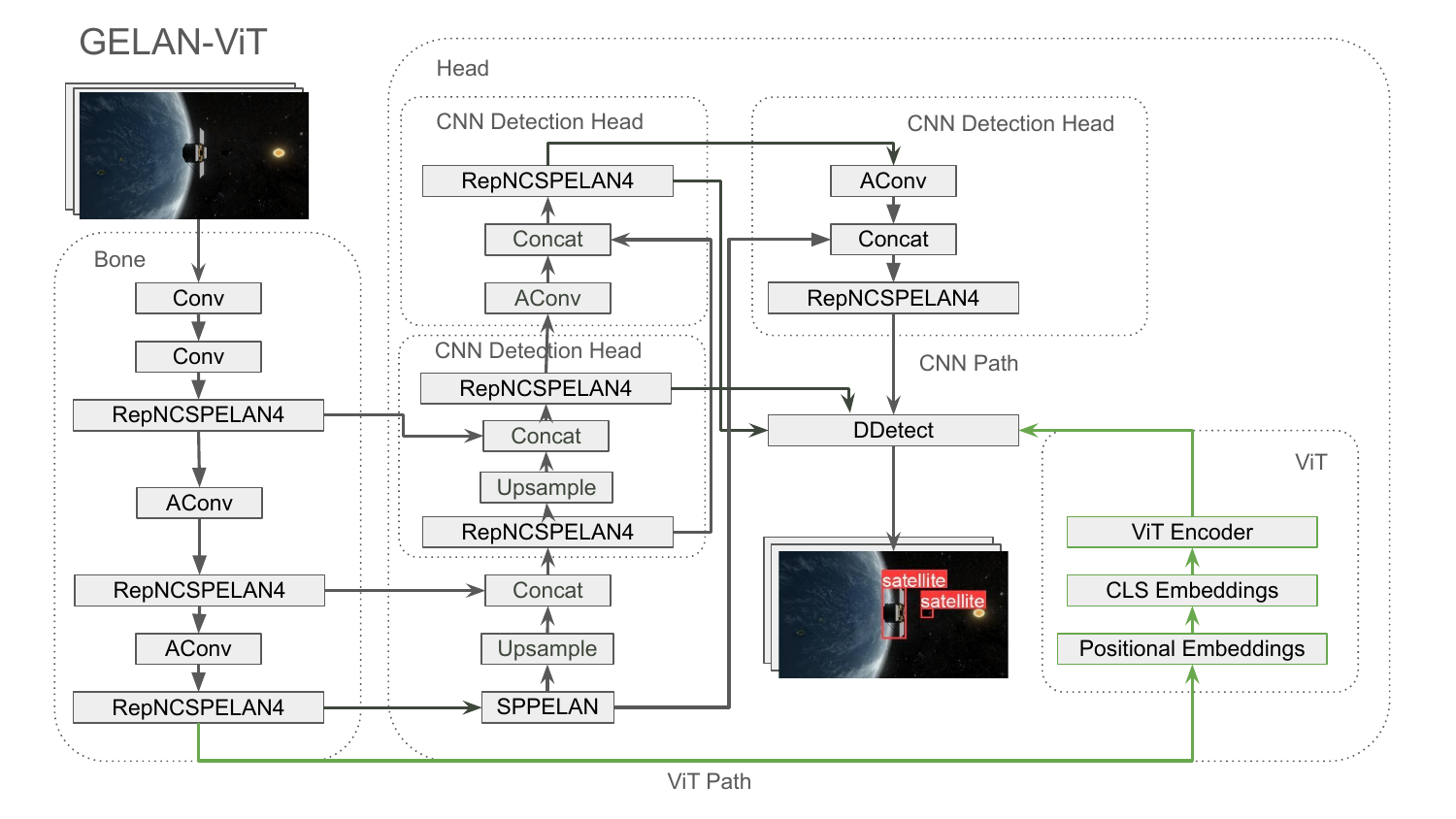}
        \caption{The architecture of the GELAN-ViT model}
        \label{fig:GELAN_ViT}
    \end{subfigure}%
    \hfill
    \begin{subfigure}[t]{0.5\textwidth}
        \centering
        \includegraphics[width=\linewidth]{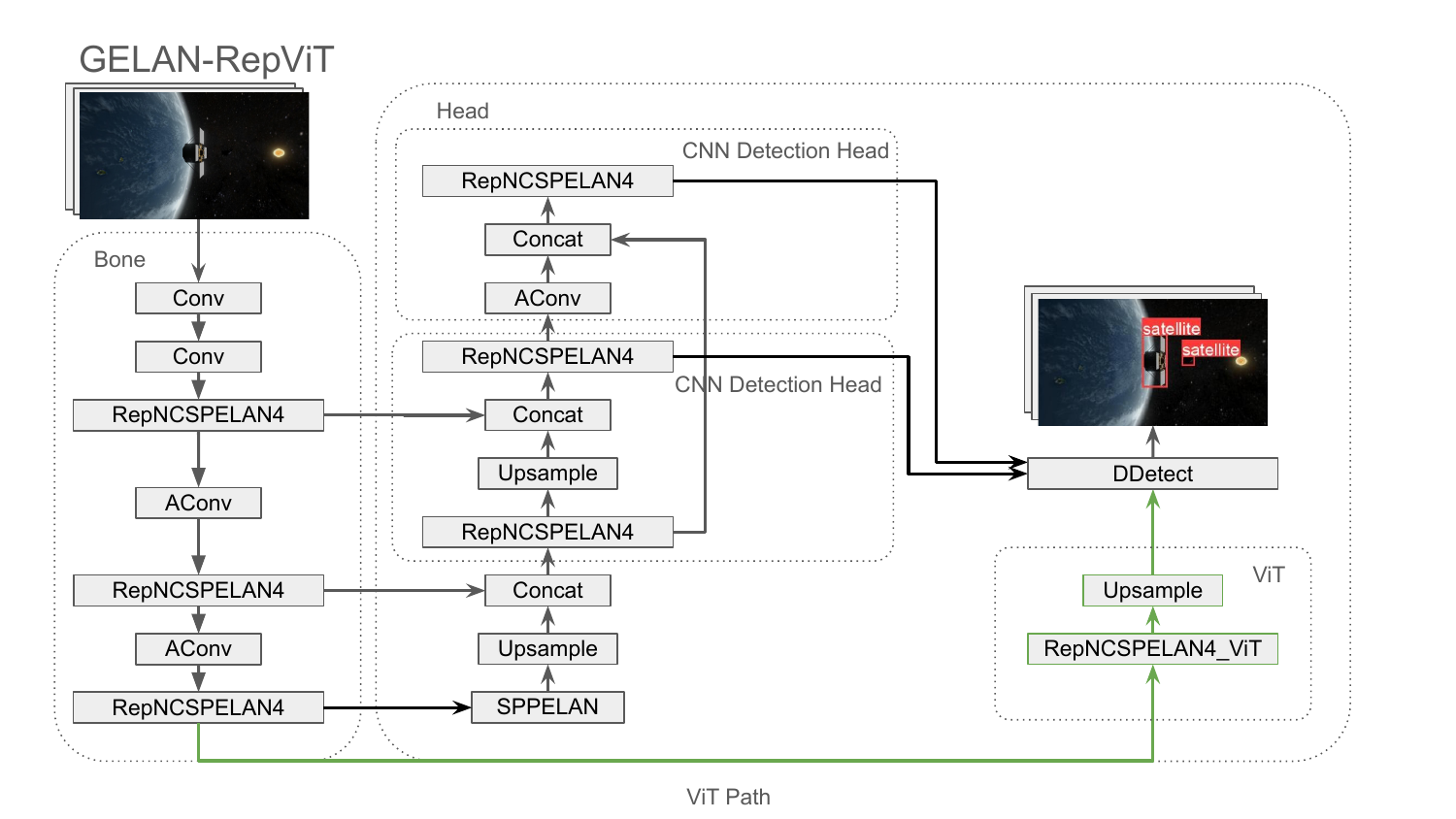}
        \caption{The architecture of the GELAN-RepViT model}
        \label{fig:GELAN_RepViT}
    \end{subfigure}
    \caption{Architectures of the proposed GELAN-ViT and GELAN-RepViT models}
    \label{fig:GELAN_models}
\end{figure*}

\subsection{Overview}

Our work focuses on enhancing the performance of GELAN-t by addressing the inefficiencies caused by combining local features ($F_l$) and global features ($F_g$) within the same pathway. Previous work has shown that when $F_l$ and $F_g$ are processed together, they compete for limited neural resources, decreasing performance. This issue arises from the \textit{information bottleneck} principle, which states that neural networks have a finite capacity for transmitting information \cite{ShwartzZiv2017OpeningTB}. When both feature types are forced to share the same pathway, the network’s capacity must be divided between them, limiting the amount of information that can be retained for both local and global features as shown in (\ref{Eq:1}):

\begin{equation}
C_{\text{total}} = C(F_l) + C(F_g)
\label{Eq:1}
\end{equation}

\begin{itemize}
    \item \(C(F_l)\) represents the capacity allocated to local features,
    \item \(C(F_g)\) represents the capacity allocated to global features.
\end{itemize}

In this setup, the network may have trouble balancing \(F_l\) and \(F_g\), leading to inefficient learning and potential information loss. This bottleneck can limit the model's ability to optimize both feature types. To address this, we separate the local and global feature extraction processes into distinct paths, allowing each to specialize in learning its feature type without interference. This separation ensures that local and global features do not compete for the same neural resources, which increases the model’s capacity to process and retain more information as shown in (\ref{Eq:2}):

\begin{equation}
C_{\text{local}} = C(F_l), \quad C_{\text{global}} = C(F_g)
\label{Eq:2}
\end{equation}

By removing the competition between local and global features, the model can retain and utilize more information from both feature types, allowing for more detailed and contextually informed predictions. We hypothesize that this separation increases the model's ability to retain critical information, resulting in better object detection performance.

\subsection{Model Architecture}

To address the issues caused by combining local and global feature extraction in the same pathway, we propose two novel architectures: GELAN-ViT and GELAN-RepViT. These models build upon the GELAN-s backbone, introducing a ViT to capture global features while maintaining the CNN-based architecture for local feature extraction. Separating these processes allows each path to specialize in its respective feature type, enabling the model to handle complex object detection tasks more effectively by focusing on local details and global context without interference.

GELAN-ViT incorporates a full ViT to capture global features with higher precision, which increases the computational cost. GELAN-RepViT simplifies the architecture, integrates a lightweight ViT encoder and reduces redundancy to optimize computational efficiency. 

\subsubsection{GELAN-ViT}
The GELAN-ViT architecture is in Fig. \ref{fig:GELAN_models}(a), which is built on top of the GELAN-s model. In this architecture, the number of neurons in each layer is scaled down by a factor of 0.25 to reduce computational complexity. It integrates core ViT components, including classify (CLS) embeddings, positional encodings, and a transformer encoder. The CLS embeddings and positional encodings provide the model with information about the spatial relationships between objects. At the same time, the transformer encoder processes the output from the last layer of the GELAN-t backbone.

As shown in Fig. \ref{fig:GELAN_models}(a), the ViT path runs parallel to the original CNN-based local feature extraction path. The outputs of the CNN and ViT paths are fed into the detection layers. The model retains the GELAN-s backbone and detection head while integrating the ViT to enhance performance.

\subsubsection{GELAN-RepViT}
The GELAN-RepViT architecture, shown in Fig. \ref{fig:GELAN_models}(b), is an extension of the GELAN-s model with three modifications. First, similar to GELAN-ViT, the number of neurons in each layer is scaled down by a factor of 0.25. Second, the architecture is simplified by removing a redundant detection head to reduce complexity. Third, we add a dedicated branch that integrates RepNCSPELAN4\_ViT, a version of RepNCSPELAN4 integrated with ViT encoder layers, as proposed by \cite{aubard2024knowledge}.

In Fig. \ref{fig:GELAN_models}(b), the RepNCSPELAN4\_ViT block operates in parallel with the local feature extraction path. The global features extracted by the ViT path are combined with the CNN features using YOLOv9’s DDetect layer. The ViT path’s output is passed through additional Upsample layers before being combined with the outputs from the CNN-based paths.

\section{Experiments}

In the experiments, we compared our proposed model with GELAN-t \cite{wang2024yolov9learningwantlearn}, YOLOv9-t \cite{wang2024yolov9learningwantlearn}, YOLOv7-t \cite{wang2023yolov7}, YOLOX-s \cite{ge2021yoloxexceedingyoloseries}, YOLOX-ViT-s \cite{aubard2024knowledge}, and MobileViT \cite{mehta2022mobilevit} using the SOD and VOC 2012 datasets. We utilized an HPC cluster with an NVIDIA V100 16 GB for training and testing. The results are averaged over 20 test runs with an image size of 640 $\times$ 640 for the SOD dataset and 320 $\times$ 320 for the VOC 2012 dataset. The current SOD dataset is designed for SOD tasks for LEO satellites and consists of satellite object classes and celestial bodies. In contrast, the VOC 2012 dataset is a commonly used benchmark dataset for object detection tasks, which contains 20 different object classes. The VOC 2012 dataset is more complex due to the variability of the classes.

Table 1 presents a comparative analysis of our proposed GELAN-ViT and GELAN-RepViT models against state-of-the-art object detection models across the SOD and VOC 2012 datasets. In general, our models achieve results comparable to the best-performing models in terms of mean average precision (mAP), specifically, mAP at 50\% IoU (mAP50) and mAP ranging from 50\% to 95\% IoU (mAP50:95). Both our models demonstrating a notable reduction in computational complexity, in terms of giga-floating point operations (GFLOPs). GELAN-ViT achieves superior performance on the SOD dataset, while GELAN-RepViT outperforms on the VOC 2012 dataset.

\begin{table*}[ht]
\centering
\caption{Comparison of different models on SODD and VOC 2012 datasets.}
\begin{tabular}{|l|c|c|c|c|c|}
\hline
\textbf{Model} & \textbf{Dataset} & \textbf{mAP50 at 1000 Epochs} & \textbf{mAP50:95 at 1000 Epochs} & \textbf{GFLOPs} & \textbf{Parameters} \\ \hline
GELAN-t \cite{wang2024yolov9learningwantlearn} & SODD & 0.944 & 0.802 & 7.3 & \textbf{1,878,819} \\ \hline
YOLOv9-t \cite{wang2024yolov9learningwantlearn} & SODD & \textbf{0.951} & \textbf{0.830} & 10.7 & 2,659,286 \\ \hline
GELAN-ViT & SODD & 0.947 & \textbf{0.815} & \textbf{5.7} & 7,327,012 \\ \hline
GELAN-RepViT & SODD & \textbf{0.950} & \textbf{0.836} & \textbf{5.2} & \textbf{1,247,811} \\ \hline
YOLOv7-t \cite{wang2023yolov7} & SODD & \textbf{0.956} & 0.759 & 13.0 & 6,007,596 \\ \hline
YOLOX-s \cite{ge2021yoloxexceedingyoloseries} & SODD & 0.858 & 0.584 & 15.2 & 5,032,866 \\ \hline
YOLOX-ViT-s \cite{aubard2024knowledge} & SODD & 0.847 & 0.575 & 15.1 & 4,884,834 \\ \hline
MobileViT \cite{mehta2022mobilevit} (width\_multiplier = 0.5) & SODD & 0.624 & 0.449 & \textbf{6.3} & \textbf{1,189,706} \\ \hline
GELAN-t \cite{wang2024yolov9learningwantlearn} & VOC 2012 & 0.605 & 0.440 & 7.4 & \textbf{1,917,148} \\ \hline
YOLOv9-t \cite{wang2024yolov9learningwantlearn} & VOC 2012 & \textbf{0.606} & \textbf{0.446} & 11.1 & 2,666,696 \\ \hline
GELAN-ViT & VOC 2012 & \textbf{0.619} & \textbf{0.460} & \textbf{5.9} & 7,350,024 \\ \hline
GELAN-RepViT & VOC 2012 & \textbf{0.607} & \textbf{0.441} & \textbf{5.7} & \textbf{1,268,180} \\ \hline
YOLOv7-t \cite{wang2023yolov7}& VOC 2012 & 0.589 & 0.390 & 13.3 & 6,066,402 \\ \hline
YOLOX-s \cite{ge2021yoloxexceedingyoloseries} & VOC 2012 & 0.585 & 0.351 & 15.3 & 5,038,395 \\ \hline
YOLOX-ViT-s \cite{aubard2024knowledge} & VOC 2012 & 0.565 & 0.340 & 15.1 & 4,890,363 \\ \hline
MobileViT \cite{mehta2022mobilevit} (width\_multiplier = 0.5) & VOC 2012 & 0.535 & 0.374 & \textbf{6.3} & \textbf{1,194,589} \\ \hline
\end{tabular}
\label{tab:enhanced_yolov9_combined}
\end{table*}

\subsection{Evaluation Metrics}

Our evaluation strategy focuses on key metrics in object detection tasks with regard to Intersection over Union (IoU), specifically, mAP50 and mAP50:95. The metrics we use are defined below.

\begin{itemize}

    \item \textbf{Average Precision (\(AP_u\)):} This is calculated as the area under the precision-recall curve for the IoU threshold \(u\) (\(p_u(r)\)).

    \[
    AP_u = \int_0^1 p_u(r) \, dr
    \]

    \item \textbf{mAP50:} This is the average of \(AP_{50}\) across all the classes \(N\):
    \[
    mAP50 = \frac{1}{N} \sum_{i=1}^{N} AP_{50,i}
    \]

    where \( AP_{50,i}\) represents the \( AP_{50}\) for class \(i\).
    
    \item \textbf{mAP50:95:} This value is averaged across all the classes \(N\):
    \[
    mAP50:95 = \frac{1}{N} \sum_{i=1}^{N} \left(\frac{1}{10} \sum_{j=1}^{10} AP_{50+5(j-1),i}\right)
    \]
    where \(j\) represents the IoU thresholds from 50\% to 95\%, and 
    \( AP_{50,i}\) represents the \( AP_{50}\) for class \(i\).
    
\end{itemize}

\subsection{Hyperparameters}

Some hyperparameters were adjusted for the training of the evaluated models. These adjustments were made manually to achieve faster convergence and obtain meaningful results within 1000 epochs without advanced hyperparameter optimization techniques.

\begin{itemize}
    \item \textbf{Learning Rate (Initial, Final):} The initial learning rate (\texttt{lr0}) was set at 0.01, with a final learning rate (\texttt{lrf}) of 0.1 using the OneCycleLR policy.
    
    \item \textbf{Loss Gains:}
    \begin{itemize}
        \item \textbf{Box Loss Gain (\texttt{box}):} Set to 0.05, this parameter controls the weight of the bounding box regression loss.
        \item \textbf{Classification Loss Gain (\texttt{cls}):} Set to 0.5, this parameter controls the importance placed on the accuracy of classifying objects within the bounding boxes.
        \item \textbf{Objectness Loss Gain (\texttt{obj}):} Set to 1.0, this parameter controls the emphasis on detecting the presence of objects within predicted bounding boxes.
    \end{itemize}
    
    \item \textbf{Data Augmentation:} 
    \begin{itemize}
        \item \textbf{HSV-Saturation (\texttt{hsv\_s}):} Set to 0.7, this parameter controls the variation in color saturation within training images.
        \item \textbf{HSV-Value (\texttt{hsv\_v}):} Set to 0.4, this parameter controls the brightness of images.
        \item \textbf{Scale Augmentation (\texttt{scale}):} Set to 0.5, this parameter controls the scaling of images during training.
        \item \textbf{Mixup (\texttt{mixup}):} Disabled (set to 0.00), this parameter controls the blending of multiple images during training, which was turned off to avoid introducing unnecessary complexity.
    \end{itemize}
    
    \item \textbf{Anchor-Multiple Threshold (\texttt{anchor\_t}):} Set to 4.0, this threshold determines the conditions under which anchor boxes are selected.
\end{itemize}

Other hyperparameters were kept at the model's default values from the model repository. These defaults were selected based on standard practices.

\subsection{Performance on SOD Dataset}
The performance of our models in the SOD dataset (SODD) is summarized in Table \ref{tab:enhanced_yolov9_combined}. Our proposed models, GELAN-ViT and GELAN-RepViT, are compared with GELAN-t, YOLOv9-t, YOLOv7-t, YOLOX-s, YOLOX-ViT-s, and MobileViT. The metrics used for the evaluation include mAP50 at 1000 epochs, mAP50:95 at 1000 epochs, GFLOPs, and the number of parameters. In calculating GFLOPs, YOLOv9 and YOLOv9-t default to an input size of 640 $\times$ 640, while MobileViT, YOLOX, and YOLOX-ViT use their respective input sizes. To ensure consistency in our comparison, we fixed the GFLOPs calculation to an input size of 640 $\times$ 640 for all models.

GELAN-ViT achieves an mAP50 of 0.947 at 1000 epochs, while GELAN-RepViT improves slightly to 0.950. Both models show a difference of less than 0.01 in mAP50 compared to GELAN-t with 0.944, YOLOv7-t with 0.956, and YOLOv9-t with 0.951. Models with significantly lower mAP50 include YOLOX with 0.858, YOLOX-ViT with 0.847, and MobileViT with 0.624.

Regarding mAP50:95, GELAN-ViT achieves 0.815 at 1000 epochs, while GELAN-RepViT improves to 0.836. Compared to GELAN-t, which achieves 0.802, and YOLOv9-t, which achieves 0.830. Models with significantly lower mAP50:95 scores include YOLOv7-t with 0.759, YOLOX with 0.584, YOLOX-ViT with 0.575, and MobileViT with 0.449. Although YOLOv7-t achieves a high mAP50, it performs significantly lower in mAP50:95. YOLOX-s, YOLOX-ViT-s, and MobileViT maintain lower performance compared to YOLOv9 and GELAN-t.

Both GELAN-ViT and GELAN-RepViT deliver comparable performance to YOLOv9 and GELAN-t while maintaining lower GFLOPs. Compared to GELAN-t, GELAN-ViT reduces GFLOPs by 1.6, and GELAN-RepViT by 2.1. Compared to YOLOv9-t, reductions are 5.0 and 5.5, respectively.

\subsection{Performance on VOC 2012 Dataset}
The performance of our models in the VOC 2012 dataset is also summarized in Table \ref{tab:enhanced_yolov9_combined}. The metrics used for evaluation include mAP50 at 1000 epochs, mAP50:95 at 1000 epochs, GFLOPs, and the number of parameters. The input size for the VOC dataset is standardized to 320 $\times$ 320. However, for consistency in GFLOPs calculation across all models, we fixed the GFLOPs calculation to an input size of 640 $\times$ 640, regardless of the original input size used in training.

In the VOC 2012 dataset, GELAN-ViT achieves the highest mAP50 of 0.619 at 1000 epochs, followed by GELAN-RepViT with 0.607. Both models outperform GELAN-t's 0.605 and YOLOv9-t's 0.606. GELAN-ViT represents a 0.014 increase over GELAN-t and a 0.013 increase over YOLOv9-t. GELAN-RepViT's mAP50 of 0.607 shows a 0.002 increase compared to GELAN-t and a 0.001 increase compared to YOLOv9-t. Models with lower mAP50 include YOLOv7-t with 0.589, YOLOX with 0.585, YOLOX-ViT with 0.565, and MobileViT with 0.535.

Regarding mAP50:95, GELAN-ViT leads with 0.460, while GELAN-RepViT achieves 0.441. In comparison, GELAN-t records an mAP50:95 of 0.440, and YOLOv9-t achieves 0.446. GELAN-ViT outperforms GELAN-t by 0.06 and YOLOv9-t by 0.054. GELAN-RepViT's mAP50:95 is 0.01 higher than GELAN-t but 0.05 lower than YOLOv9-t. Models with lower mAP50:95 scores include YOLOv7-t with 0.390, YOLOX with 0.351, YOLOX-ViT with 0.340, and MobileViT with 0.374.

While GELAN-ViT achieves superior performance and GELAN-RepViT delivers performance comparable to YOLOv9-t and GELAN-t, both models maintain low GFLOPs. Compared to GELAN-t, GELAN-ViT reduces GFLOPs by 1.5 and GELAN-RepViT by 1.7. Compared to YOLOv9-t, GELAN-ViT and GELAN-RepViT show reductions of 5.2 and 5.4 GFLOPs, respectively. In contrast, YOLOv7-t, YOLOX-s, YOLOX-ViT-s, and MobileViT achieve a substantially lower performance than YOLOv9-t and GELAN-t.

\subsection{Model Stability}
In the SODD, YOLOv7-t achieves the mAP50 in the first place of 0.956, surpassing YOLOv9-t, which scores 0.951. However, its mAP50:95 is significantly lower at 0.759 compared to YOLOv9-t's 0.830. This indicates that while YOLOv7-t shows strong performance in mAP50, it exhibits inconsistent precision across different IoU thresholds and struggles to achieve higher accuracy at these more demanding levels.

Although YOLOv7-t's high mAP50 in the simpler SODD highlights its capability in less complex scenarios, this advantage diminishes in the more complex VOC 2012 dataset. In the VOC 2012 dataset, YOLOv7-t's performance is lower than that of YOLOv9-t, with an mAP50 of 0.589 compared to YOLOv9-t's 0.606 and an mAP50:95 of 0.390 compared to YOLOv9-t's 0.446.

In general, GELAN-based models, including GELAN-t, YOLOv9-t, GELAN-ViT, and GELAN-RepViT, perform better in complex scenarios and demonstrate more stable and consistent performance across various IoU thresholds in both datasets, particularly maintaining higher mAP50:95 values.

GELAN-ViT, with its higher parameter count and relatively complex architecture, is expected to perform better on complex datasets due to its ability to capture more intricate patterns. However, this complexity also increases the risk of overfitting when applied to simpler datasets. In contrast, GELAN-RepViT, a more streamlined model with fewer parameters, is better suited for simpler datasets, demonstrating its efficiency with slightly higher performance.

The significant increase in parameters for GELAN-ViT is largely due to the integration of full ViT layers, which tend to have a high parameter count and high computational complexity. However, since the ViT layers are added after the detection head, the input size to the ViT has already been reduced, resulting a fewer computational operations, which allows the model to maintain lower GFLOPs despite a large number of parameters.

\subsection{Discussion of Model Superiority}
ViTs utilize the self-attention mechanism to learn global features by capturing long-range dependencies within an image, which is crucial for complex object detection tasks requiring scene understanding and identifying object interactions.

On the other hand, CNNs excel at extracting detailed features through local patterns. They are particularly effective in recognizing edges, textures, and shapes, essential for accurately identifying individual objects within an image. By focusing on refining these local details, CNNs construct precise representations of objects, ensuring that their specific characteristics are captured effectively.

Our model optimizes feature extraction by separating the ViT and CNN pathways. This allows CNNs to specialize in local feature extraction, while ViTs focus on capturing the global context. By separating these paths, our model prevents interference between the local and global feature learning processes, allowing each network to specialize without distraction. This design also addresses the information bottleneck problem: when local and global features share the same pathway, information is lost due to the limited capacity of the neurons. By using separate pathways, each type of feature is retained more effectively. Later in the network, these features are fused, combining local and global strengths to improve overall performance.

In comparison, YOLOX-ViT's performance decreases due to the competition between local and global feature extraction within the same pathway. This shared processing introduces an information bottleneck, limiting the model's ability to capture both feature types effectively. A similar observation was made by Aubard \textit{et al.} \cite{aubard2024knowledge}, where integrating ViT increased false positives and contributed to potential overfitting. These findings highlight a trade-off between the benefits of enhanced global feature extraction and the efficiency required for accurate object detection, as reflected in YOLOX-ViT's performance.

A similar pattern is observed in MobileViT, but since its architecture is designed to be lightweight with fewer parameters and narrower layers, the impact of the information bottleneck becomes even more noticeable. MobileViT’s reduced capacity exacerbates the challenge of balancing \(F_l\) and \(F_g\), leading to greater information loss and a more significant decline in accuracy compared to larger models like YOLOX-ViT or GELAN-ViT.

\section{Conclusion}

In this paper, we proposed two novel models, GELAN-ViT and GELAN-RepViT, for SOD tasks. The proposed models combine the strengths of CNN and ViT architectures and demonstrate competitive accuracy on the VOC 2012 dataset and SOD datasets while maintaining lower computational costs compared to the latest DL models. Although the proposed models have shown superior performance metrics that are suitable for SOD tasks, the high number of parameters of GELAN-ViT shows room for improvement. Our future work will focus on streamlining the architecture to reduce parameters while preserving the same performance, which can make the model require less memory on the target satellite platforms.


%

\section*{Acknowledgment}
We acknowledge the support of the Natural Sciences and Engineering Research Council of Canada (NSERC), [funding reference number RGPIN-2022-03364].

\ifCLASSOPTIONcaptionsoff
  \newpage
\fi



\bibliographystyle{IEEEtran}
\bibliography{IEEEabrv,references}
%



%

\begin{IEEEbiography}{Michael Shell}
Biography text here.
\end{IEEEbiography}

\begin{IEEEbiographynophoto}{John Doe}
Biography text here.
\end{IEEEbiographynophoto}


\begin{IEEEbiographynophoto}{Jane Doe}
Biography text here.
\end{IEEEbiographynophoto}




\end{document}